\documentclass[conference]{IEEEtran}
\usepackage[utf8]{inputenc} 
\usepackage[T1]{fontenc}    
\usepackage{url}            
\usepackage{amsfonts}       
\usepackage{nicefrac}       
\usepackage{microtype}      
\usepackage{graphicx}
\usepackage{subfig}
\usepackage{amsmath}
\ifCLASSINFOpdf
\else
\fi

\begin{document}
\IEEEoverridecommandlockouts
\IEEEpubid{\makebox[\columnwidth]{978-1-6654-4067-7/21/\$31.00 \copyright 2021 IEEE \hfill} \hspace{\columnsep}\makebox[\columnwidth]{ }}
%
\title{Resilience of Autonomous Vehicle Object Category Detection to Universal Adversarial Perturbations}

\author{\IEEEauthorblockN{Mohammad Nayeem Teli}
\IEEEauthorblockA{Department of Computer Science\\
University of Maryland\\
College Park, Maryland 20740\\
}
\and
\IEEEauthorblockN{Seungwon Oh}
\IEEEauthorblockA{Department of Electrical and Computer Engineering\\
University of Maryland\\
College Park, MD 20740 \\
}
}


%


\maketitle

\begin{abstract}
Due to the vulnerability of deep neural networks to adversarial examples, numerous works on adversarial attacks and defenses have been burgeoning over the past several years. However, there seem to be some conventional views regarding adversarial attacks and object detection approaches that most researchers take for granted. In this work, we bring a fresh perspective on those procedures by evaluating the impact of universal perturbations on object detection at a class-level. We apply it to a carefully curated dataset related to autonomous driving. We use Faster-RCNN object detector on images of five different categories: person, car, truck, stop sign and traffic light from the COCO dataset, while carefully perturbing the images using Universal Dense Object Suppression algorithm. Our results indicate that person, car, traffic light, truck and stop sign are resilient in that order (most to least) to universal perturbations. To the best of our knowledge, this is the first time such a ranking has been established which is significant for the security of the datasets pertaining to autonomous vehicles and object detection in general. 
\end{abstract}


%
\IEEEpeerreviewmaketitle

\section{Introduction}
In an autonomous vehicle set up we can categorize the interactions between various components based on the interactions of a vehicle and one or more of the following: landscape (e.g., roads, barriers, buildings, trees etc.), other automobiles, people, and road signs. Autonomous vehicles often recognize each of these categories through the use of Machine Learning models for object detection. While objection detection techniques have made a huge progress, so have the various adversarial attacks, as explained in the next few sections. Most of the studies have heavily concentrated on the techniques of adversarial attacks and the means of making them more and more efficient in breaking the object detector. A lot of research has also concentrated on defense approaches. However, there is very limited or no research done to empirically understand robustness of each individual object category. In this work, we mostly concentrate on specific object categories and their ability to withstand adversarial attacks. 

More specifically, we answer questions like, is it more difficult to recognize a person in comparison to a car when an object detector is subjected to universal perturbations?
It is very important to understand the impact of such adversarial attacks at a category level so that targeted efforts could be made to address such weaknesses in object detection. Also, different object categories have different features and understanding the affects on them individually would enable us to build systems that can encompass methods to overcome the weaknesses of an object detector.

Due to the vulnerability of deep neural networks to adversarial examples, adversarial attacks have been receiving huge attention over the past several years~\cite{biggio2018wild}. In the context of the adversarial attacks, universal perturbations research has also received a lot of interest~\cite{univadpersurvey} since it is much more efficient and transferable than image-specific attacks. However, most of it is directed toward image classification and not much toward object detection~\cite{universaladvpert}. More importantly, we observed that there are two conventions that most work on adversarial attacks seem to be holding on: evaluation of adversarial attacks on all (1) \textbf{classes} and (2) \textbf{images}. 

In this paper, we address issues pertaining to the impact of dataset curation and its need, and, the effect of universal perturbations on individual categories of objects. 
First of all, we question whether the experiments on adversarial attacks have been measuring their impact rigorously, for instance, quantified by the fooling rate (the proportion of images that change labels when perturbed by our universal perturbation). Many experiments have been naively using the whole dataset which may lead to assessing the effect of an attack in a more optimistic way. It is owing to the fact that the object detector may already have some incorrect predictions on some of the images in the first place, regardless of the perturbation. The number of those images could be quite high. This is supported by the fact that the mAPs on many datasets are still not high enough to ensure this even on clean images, for example, the COCO AP for the state-of-the-art object detector is still less than 60. Thus, we argue that those examples should not be counted as an example of a successful attack and also be removed from the dataset to prevent dataset bias from the evaluation.

Secondly, we also observed that there is less research toward understanding the impact of universal perturbations on specific categories related to autonomous driving. Most of the experiments and evaluations are conducted while including all the classes in a dataset. This may widen the gap between understanding the effect of universal perturbations in general. In more specific applications a successful adversarial attack may perform differently in the self-driving car domain. Given that deep learning has various applications beyond just self-driving cars, we need to also consider evaluating adversarial attacks at a class-specific level and compare which attack is more deadly to a specific problem,  and not just in general. To those who are closely related to the domain, this could have more meaningful implications rather than a general evaluation on a general-purpose dataset. In the context of autonomous vehicles, we would like our system to be more reliable, and robust to attacks on the person class than may be those on stop signs. This is due to the fact that the consequences for a vehicle to crash into a person in comparison to a stop sign is vastly different.

In this work, we incorporate dataset curation and class-level evaluation of robustness into experimental framework. We present a rigorous evaluation on the impact of universal perturbations on detecting specifically five object categories related to autonomous vehicles: person, stop sign, car, truck and traffic light. We run a series of controlled experiments by perturbing images of these categories using universal perturbations. Finally, we rank the vulnerability of detecting each object category by calculating the metrics for each object category individually and compare them.

We summarize the main contributions of this paper as follows: (1) firstly, we demonstrate the vulnerability of object detectors in detecting autonomous vehicle-related categories to universal perturbations. This is achieved using a simple iterative method of using the gradient of soft-max output with respect to input image and perturbation, (2) secondly, we establish a class-level vulnerability ranking with regards to adversarial attacks, and (3) thirdly, we isolate the impact of adversarial perturbations. This is done by including only those images wherein the detector correctly predicted at least one of the designated classes.

\section{Related Works}

\subsection{Adversarial Attacks on Object Detection}
Deep Neural Networks are ubiquitous in today's world within and outside the lab that have been built to alter the way applications are run. Its impact can be seen all around us. However, it does not come without its pitfalls. They are susceptible to adversarial attacks \cite{szegedy}\cite{ozdag}\cite{akhtar}\cite{chakraborty}. This is a result of many noise injection techniques that cause failure in object detection and classification. Kurakin et al.~\cite{kurakin} specifically demonstrated the attacks on autonomous vehicles wherein the traffic signs were manipulated. 


Object detection in images, specifically related to autonomous vehicles, that is the focus of this work, come under adversarial attacks as well~\cite{serban}. Adversarial examples are generated to be similar to the actual data,  $\mathbb{D}$, with the goal of a mis-classification. Consider a classification function, $f$ that correctly classifies an image, $x$, with a label $y$, and $x'$ gets incorrectly classified with a label $\hat{y}$: $\arg\min_\eta f(x + \eta)=\hat{y}$. According to Szegedy et al.~\cite{szegedy} we can solve the following optimization problem to find the perturbation,

\begin{equation*}
    \min_{x'} ||x'-x||_p,\hspace{0.4in}
    s.t. \hspace{0.1in} f(x') = \hat{y}
\end{equation*}
 where $||.||_p$ is a distance defined on the metric space $X$. The most commonly used distance metric for object recognition in adversarial attacks is the $p$-norm:
 
\begin{equation*}
    ||x||_p = \bigg(\sum\limits_{i=1}^n|x_i|^p\bigg)^{1/p},
\end{equation*}

where $p\in \{0,2,\infty\}$. According to Serban et al.~\cite{serban} the choice of $p$ is picked as follows:
\begin{itemize}
    \item $p=0$ measures the dissimilarity in coordinates between the perturbed and original images with regards to the number of pixels altered in the original image.
    \item $p=2$ measures the Euclidean distance between the original and the adversarial image.
    \item $p=\infty$ measures the maximum bound for altering each pixel in an image without restricting the number of changed pixels.
\end{itemize}

Deng et al.~\cite{deng} presented an analysis of adversarial attacks on autonomous driving models using five adversarial attacks and four defense methods on three driving models and found them to be vulnerable.
Sharma et al.~\cite{sharma} demonstrated the vulnerability of the machine learning models of autonomous vehicles in their failure of detecting adversarial attacks. 
\subsection{Universal Adversarial Perturbation}
The attacks explained above are mostly image-specific, 
However, a much stronger class of adversarial attacks, called universal attack generates a single adversarial example to mis-classify all images in the dataset~\cite{universaladvpert}. Benz et al.~\cite{benz} introduced a double targeted universal adversarial perturbations (DT-UAPs) to bridge the gap between the instance discriminative image-dependent perturbations and the generic universal
perturbations and its potential as a physical attack.
Sitawarin et al.~\cite{sitawarin} expand the scope of adversarial attacks using Out-of-distribution attacks by enabling the
adversary to generate these starting from an arbitrary point in
the image space. Prior attacks are restricted to
existing training/test data (In-Distribution). They demonstrated that Out-of-Distribution attacks can outperform In-Distribution attacks on classifiers and 
exposed a new attack vector for these defenses.

Metzen et al.~\cite{metzen} proposed attacks against semantic image segmentation which removes a target class, such as, pedestrians, from the segmentation while leaving the segmentation otherwise unchanged. Li et al.~\cite{advpertod} propose, for the first time, an iterative method to generate universal adversarial perturbations against object detection. Yingwei et al.~\cite{yingwei} propose that regionally homogeneous perturbations can well transfer across
different vision tasks (attacking with the semantic segmentation task
and testing on the object detection task) and the defenses
are less robust to regionally homogeneous perturbations.

 As is noticed in the previous works the emphasis has been on the different adversarial attacks and defenses against such attacks. However, to the best of our knowledge, very little to no work has been performed to understand the robustness of the image categories. Our work focuses on the area of generating universal adversarial perturbations against object detection, but proposes a new evaluation framework.

\section{Dataset curation}
In this research, we curate image categories from the popular and challenging COCO dataset by focusing on the categories specific to autonomous vehicles. Since it is an object detection dataset, most of the images would contain more than one category in a single image. This is done by design, since in a real world scenario a given scene would contain more than a single object category. Table \ref{categories} contains the number of such instances for each class in the COCO training set. 
\begin{table}
    \caption{Autonomous vehicle object categories in COCO dataset}
    \label{categories}
    \begin{center}
    \begin{tabular}{|c|c|}
    \hline
     \textbf{category} & \textbf{Number of training samples}\\
    \hline
    person & 64115\\
    \hline
    bicycle & 3252\\
    \hline
    car & 12251\\
    \hline
    truck & 6127 \\
    \hline
    bus & 3952 \\
    \hline
    motorcycle & 3502\\
    \hline
    traffic light & 4139\\
    \hline
    stop sign & 1734\\
    \hline
    \end{tabular}
    \end{center}
\end{table}

\begin{figure*}[!t]
    \centering
   {{\includegraphics[width=5cm]{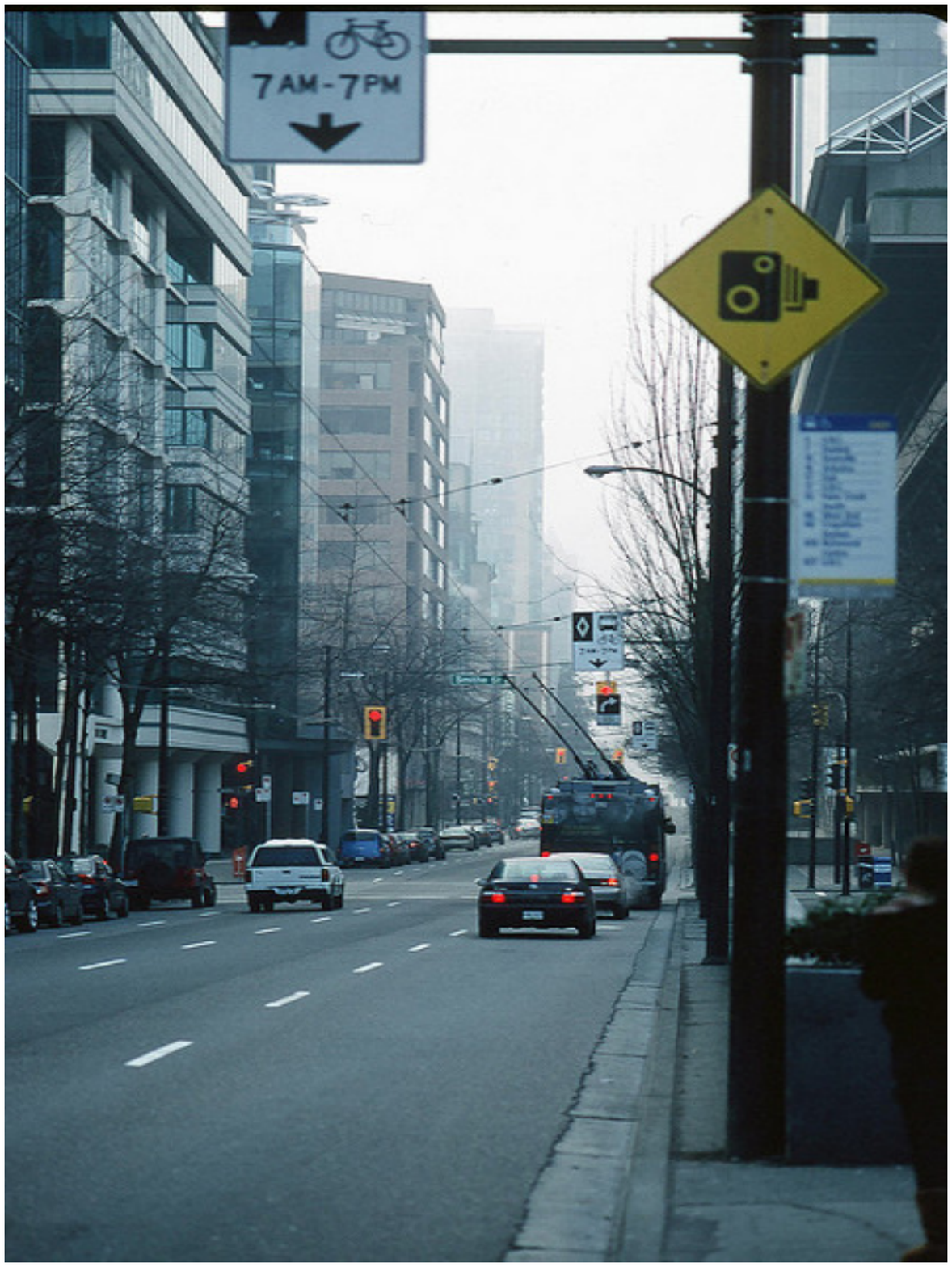}}}%
    \qquad
    {{\includegraphics[width=5cm]{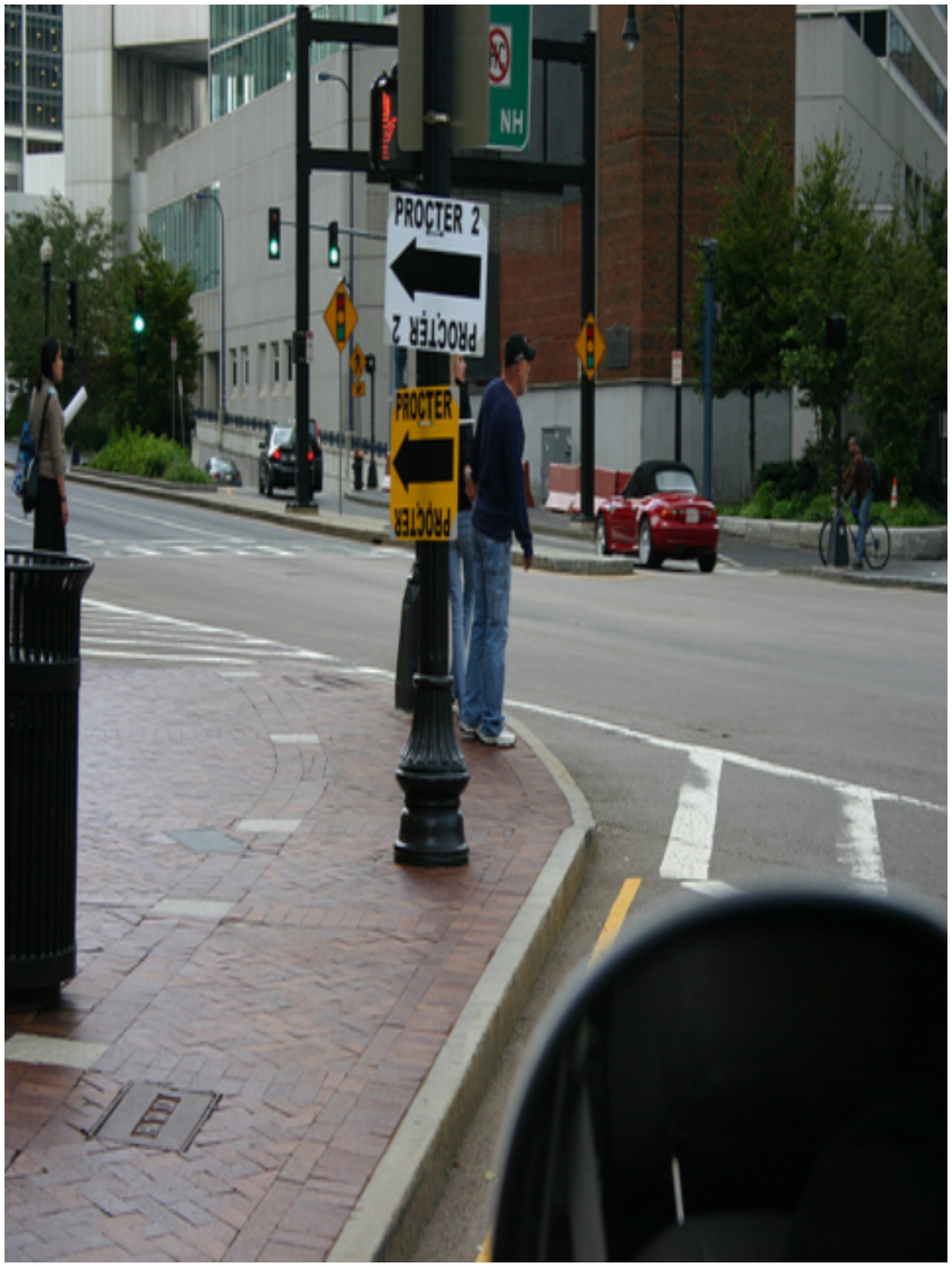}}}%
    \qquad
   {{\includegraphics[width=5cm]{figs/stop_sign.pdf}}}%
    \qquad
    {{\includegraphics[width=5cm]{figs/traffic_light.pdf}}}%
     \qquad
    {{\includegraphics[width=5cm]{figs/truck.pdf}}}%
    \caption{Sample images of the selected object categories}%
    \label{objcats}
\end{figure*}

Among the vehicle categories, we only look at the top two by the number of instances and leave out bicycle, bus, and motorcycle and finalized with 5 categories: person, car, truck, stop sign, and traffic light for this experiment. Also, we only include the training set in COCO because the number of instances of each category is large enough, and most importantly, we meticulously study the independent impact of perturbations and leave out any incorrect classifications due to the object detection failure.
To this end, using the training set on which an object detector has been trained would give us higher probability of collecting images with true positives. Some of the sample images from the dataset are shown in Figure~\ref{objcats}.

In order to rule out missed detections due to the failure of object detector, we curate the dataset based on the results of the object detector prior to perturbing an image. As long as the object detector is able to detect at least one instance of the object category, we would retain that image in our dataset. In case the detector missed all the object instances of a target class in an image, that image would not be used further. 

\section{Methods}
\subsection{Adversarial perturbation}
Universal perturbation is an image-independent perturbation that would cause a label change for a number of images~\cite{universaladvpert}. Our goal is to test the impact of such perturbations in object detection. Li et al.~\cite{advpertod} introduced Universal Dense Object Suppression (U-DOS) algorithm to find universal perturbations that would blind an object detector on training set, $I$, so that it may fail to find objects in most of the images in $I$ and at the same time remain imperceptible to a human eye. We use this approach since it is for the first time that the existence of universal perturbations on object detection tasks is performed. We made two modifications to the approach, though: 1) to simplify the loss function, we do not include the background probability in the objective function , and, 2) we use a normalized L1-norm by dividing it by the number of pixels. 

U-DOS is an iterative algorithm that updates the universal perturbation, $\textbf{v}$, by generating a vector, $v_i$, for each image, $I_i \in I$, so that the detector, $D$, fails to find any objects in the perturbed image, $I_i + \textbf{v} + v_i$. 
We modify the original objective function in Li et al.~\cite{advpertod} to only include the sum of class confidence probabilities of detection instances. The modified objective function is given as, 

\begin{equation}
\label{opteq}
\mathcal{F}(I_i,\textbf{v},v_i) =\sum_{r_n\in \mathcal{R}_{D,I_i,\textbf{v}+v_i}}P_{obj}(r_n|I_i + \textbf{v} + v_i)
\end{equation}
where, 
$P_{obj}(r_n|I_i + \textbf{v} + v_i)$ is the probability of the object class with the highest confidence, and $\mathcal{R}_{D,I_i,\textbf{v}+v_i}$, is the set of detection results on the perturbed image. Equation~\ref{opteq} is minimized while limiting the magnitude of the perturbation, $v_i$, as
\begin{equation}
v_i \leftarrow \operatorname*{argmin}_{v_i'}\bigg[\mathcal{F}(I_i, \textbf{v},v_i') + ||\textbf{v}+v_i'||\bigg]
\end{equation}
To solve for $v_i$, the algorithm starts with a zero-vector and optimizes it with gradient descent.
\begin{equation}
\begin{split}
&v_i^{(0)} = \textbf{0} \\
&v_i^{(n+1)} = v_i^{(n)} - \alpha \nabla_{v_i^{(n)}}(\mathcal{F}(I_i,\textbf{v},v_i^{(n)}) + ||\textbf{v} + v_i^{(n)}||)
\end{split}
\end{equation}

where, $\alpha$ is the learning rate and the gradient, $\nabla_{v_i^{(n)}}(*)$, is computed with back-propagation.

We modified Li et al.'s objective function to make it more applicable to object detectors in general: originally the paper included the term of increasing the background class probability and decreasing the ones for specific/all object class, but we removed the term increasing the probability of the background class because not all object detectors output a soft max probability that includes the background class.

\subsection{Metrics}
In order to evaluate the performance of the universal perturbations in blinding the object detector, we use two metrics proposed by~\cite{advpertod}: \textbf{image-level blind degree} and \textbf{instance-level blind degree}. 

\subsubsection{Image-level blind degree}
Image-level blind degree is defined as the ratio of the images in which the object detector, $D$, can find at least one object with confidence above a threshold, $\theta$, to the total number of images, $N$. It is expressed as
\begin{equation}
B_{img}(D,\textbf{v},\theta) = \frac{\sum\limits_{i=1}^N ind(I_i, D,\textbf{v},\theta)}{N}
\end{equation}
where, $N$ is the total number of images, and
$I_i$ denotes the $i$-th image, $\textbf{v}$ the universal perturbation added to these images, 
$ind(I_i, D,\textbf{v},\theta)$ is an indicator function
represented as 

\[
    ind(I_i, D, \textbf{v}, \theta) = \left\{\begin{array}{lr}
        1, & \text{if } \exists r \in \Re_{D,I_i,\textbf{v}}, P_{obj}(r|I_i + \textbf{v}) > \theta\\
        0, & \text{otherwise} 
        \end{array}
        \right.
\]
 
 where the set of detection results on the perturbed image is $D(I_i + \textbf{v})$ given by $\Re_{D,I_i,\textbf{v}} = D(I_i + \textbf{v})$, with $\textbf{v} \in \mathbb{R}^d$ being the perturbation. $P_{obj}(r_n|I_i + \textbf{v} + v_i)$ is the probability of the object class with the highest confidence and the 
 image-independent perturbation, $\textbf{v}$, must satisfy the following two constraints: 
  \begin{enumerate}
  \item $\Re_{D,I_i,\textbf{v}} = \emptyset$
  \item $||\textbf{v}||_{\infty} < \xi$
  \end{enumerate}
  
\subsubsection{Instance-level blind degree}
Instance-level blind degree calculates the average number of instances that the detector, $D$ finds in each image with confidence beyond the given threshold $\theta$ and is expressed as
 
 \begin{equation}
 \mathcal{B}_{ins}(D,\textbf{v},\theta) = \frac{\sum\limits_{i=1}^N|\{r|r \in \mathcal{R}_{D,I_i,\textbf{v}} \mbox{ and } P_{obj}(r|I_i+\textbf{v}) > \theta\}|}{N}
 \end{equation}
 
 where $|*|$ denotes the number of elements in the set.
 
 \section{Experiments}
 \label{rslts}
\begin{figure*}[!t]%
    \centering
    {{\includegraphics[height=5cm]{figs/before_after_perturbation/stopsign2_clean.png}}}%
    \qquad
    {{\includegraphics[height=5cm]{figs/before_after_perturbation/stopsign2_perturbed.png}}}%
     \qquad
    {{\includegraphics[height=5cm]{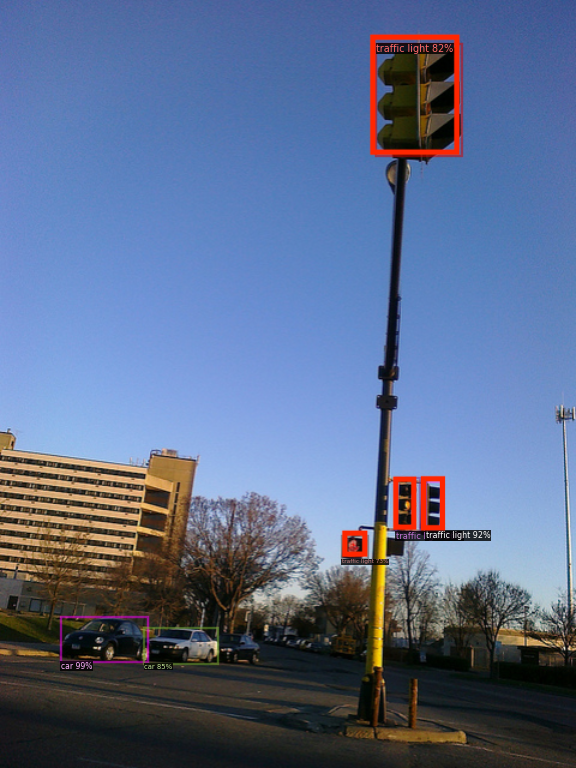}}}%
     \qquad
    {{\includegraphics[height=5cm]{figs/before_after_perturbation/trafficlight_perturbed.png}}}%
    \caption{Samples detection results with and without perturbation. 1st and 3rd columns: detection with clean images. 2nd and 4th columns: detection with perturbed images}%
    \label{pertres}%
\end{figure*}

In order to test the impact of perturbations on our selected categories, we use Faster-RCNN-FPN-Resnet50 (trained on COCO2017 train) model in Detectron2~\cite{fasterRCNN}~\cite{detectron2} library because of its high mean Average Precision (mAP) on the COCO2017 test. We also fine-tuned three hyperparameters: number of epochs($n\_epoch$), step size ($\alpha$) and $\xi$ for the max inf-norm of the perturbations. After a series of pilot experiments, we settled on the following values for the parameters: $n\_epoch=250,\xi$ (max $l_p$ norm) = 10, $\alpha = 20, max\_imgs = 500$, and $score\_threshold, \theta = 0.7$. In our experiments, we keep track of the instance-level and image-level blind degree as we vary perturbation norm for each of the five categories of objects. This is done so that the object detector fails to find objects of a given class at a certain norm. Figure~\ref{pertres} visually shows two examples of the object detector blindness before and after the perturbations.
 
\begin{figure*}[!t]%
    \centering
    \subfloat[\centering Image level blind degree per epoch]{{\includegraphics[scale=0.6]{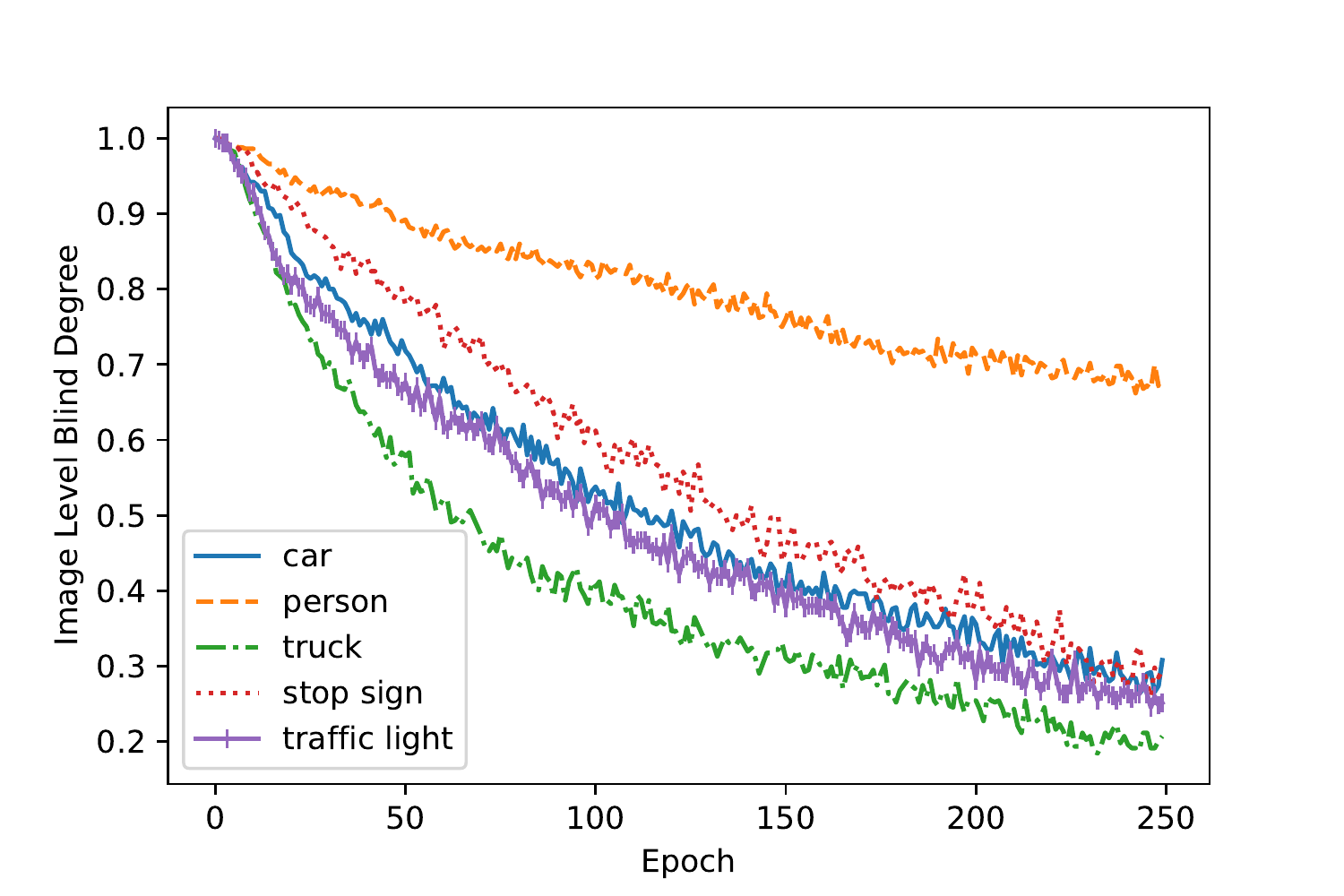}}}
    \subfloat[\centering Instance level blind degree per epoch]{{\includegraphics[scale=0.6]{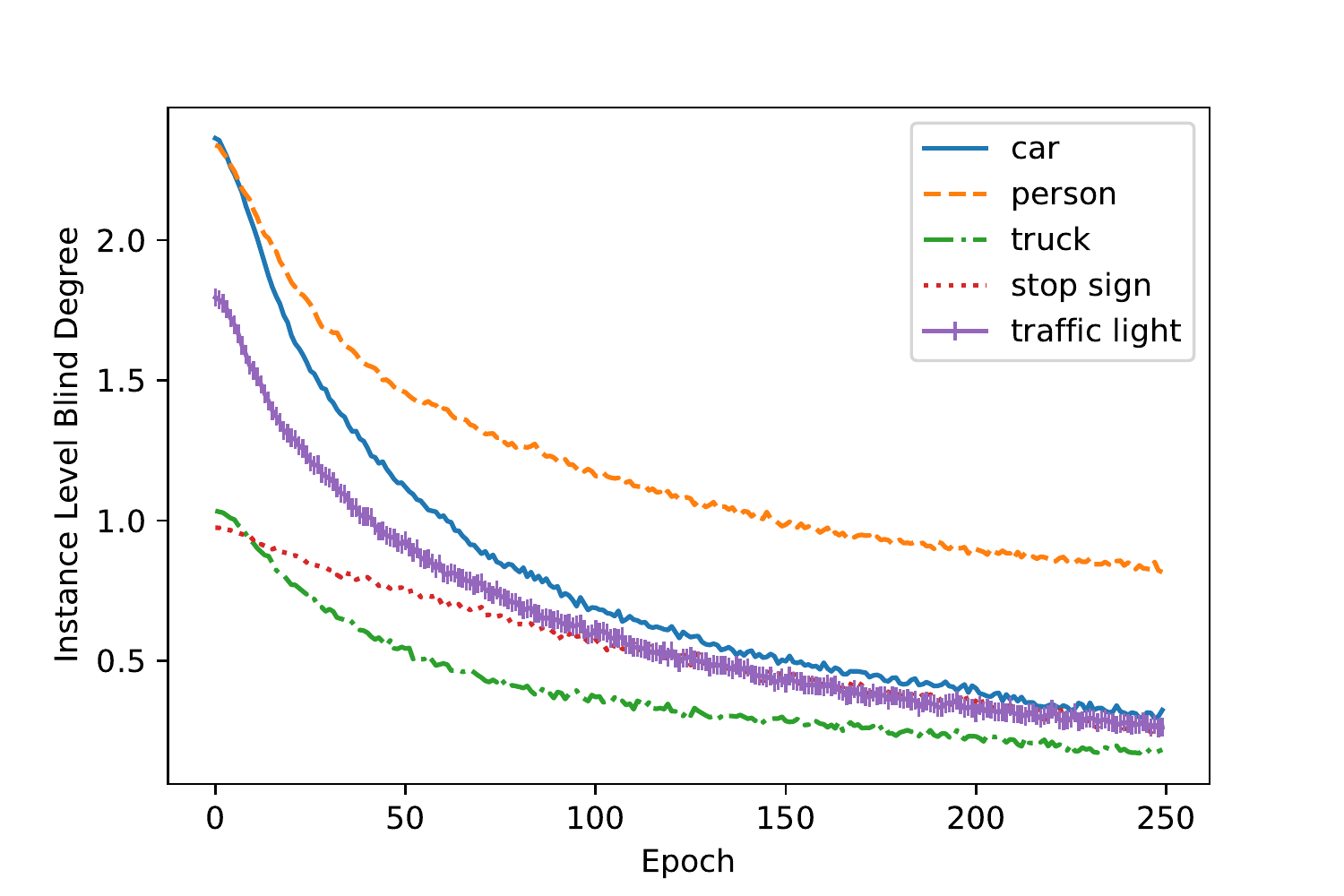}}}
     \qquad
    \subfloat[\centering Image level blind degree vs. norm]{{\includegraphics[scale=0.6]{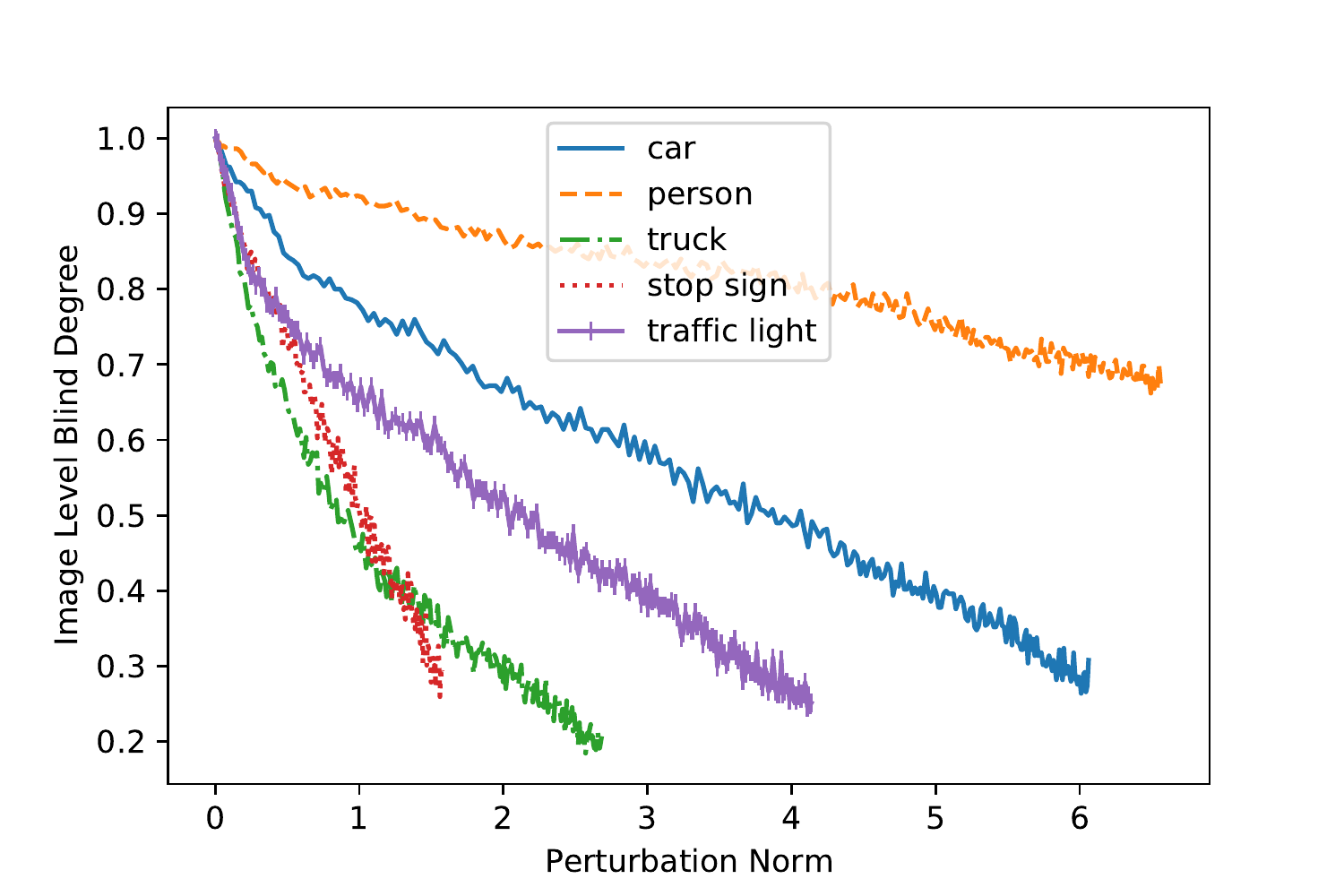}}}
    \subfloat[\centering Instance level blind degree vs. norm]{{\includegraphics[scale=0.6]{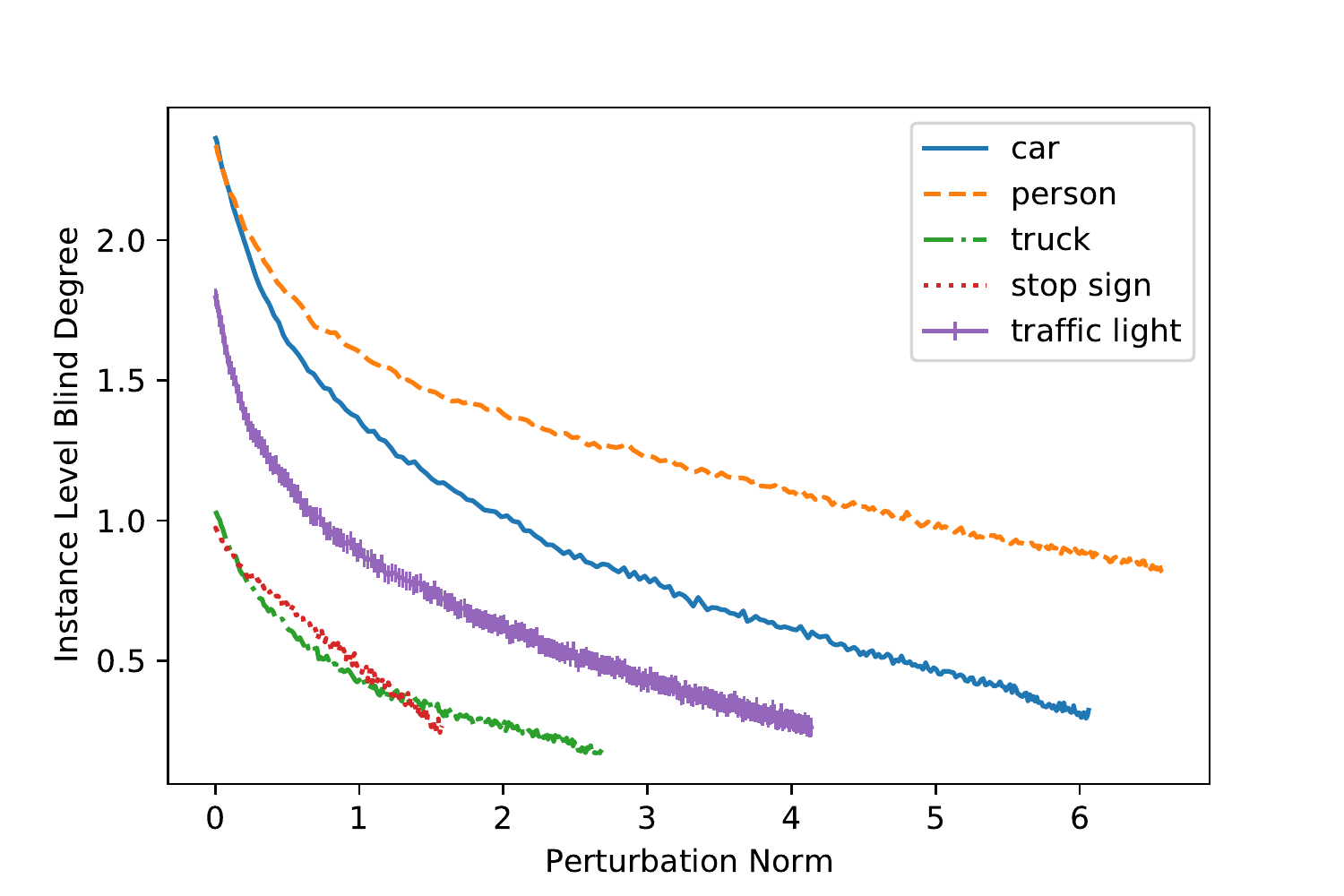}}}
    \caption{Image-level and instance-level blind degree vs. epochs and norms. Both blind degrees are decreased as we increase the norm of perturbation}%
    \label{results}%
\end{figure*}

To evaluate universal perturbations on each category, we conduct a targeted attack on each category. We compute an universal perturbation that attacks only the targeted category by considering only the object proposal classified that category and optimizing on the loss based on those object proposals to remove final detections of that class. Figure~\ref{results} presents the effect of perturbations on each category as the epochs and the norm are increased. 

Based on Figure~\ref{results}, Table~\ref{ranks} presents the ranks of resilience of each of our categories. It is evident that the person is the most resilient category at image-level as well as instance-level, followed by car, traffic light, truck and stop sign as the norm is increased. It is also clear that person category is again most resilient to perturbations as you increase the number of steps in projected gradient descent to find a stronger perturbation. However, norm is a more significant metric as it is related to the visibility of the perturbations. It is also notable that a single universal perturbation is able to make the detector fail in detecting objects of all categories, except for, the person category in most of the images, at a norm which is imperceptible to humans.

  \begin{table}[!t]
\caption{Rank of image level and instance level blind degree resilience of different categories to perturbations as epochs and norm increase }
\begin{center}
\begin{tabular}{|l|l|l|l|l|l}
    \hline
      \multicolumn{2}{|c}{\textbf{Epochs}} &
      \multicolumn{2}{|c|}{\textbf{Norm}} \\
    
    \textbf{Image}  & \textbf{Instance} & \textbf{Image} & \textbf{Instance}  \\
    \hline
   person & person & person & person  \\
    \hline
   stop sign & car & car & car  \\
    \hline
    car & traffic light & traffic light & traffic light  \\
    \hline
     traffic light & stop sign & truck & truck  \\
    \hline
     truck & truck & stop sign & stop sign  \\
    \hline
  \end{tabular}
\end{center}
\label{ranks}
\end{table}%

\section{Conclusion}
In this paper, we explore the vulnerability of object detection on five autonomous vehicle object categories to universal perturbations. In diligently designed experiments, we carefully curated the realistic and diverse COCO dataset and isolate the failure of object detection without perturbations, from the failure with perturbations. This work lies at the core of the security of datasets in the realm of autonomous vehicle driving. To the best of our knowledge, this is the first study of its kind that ranks the resilience of these five categories to universal perturbations. These results are significant since the perturbations can be imperceptible to humans while causing real harm in an adversarial attack. We believe a more fine-grained and detailed view on the experiments can be beneficial to the rigorous and evaluation of adversarial attacks. 
\section{Future Work}
Having established a categorical order based on the resilience of image categories to adversarial attacks, this research could take many different directions. One of the possible future effort would be to build and train object detectors while incorporating special features of each of the categories. Another possible direction would be to study the reasons behind the resilience of each of these categories and employ that in training robust object detectors. A different approach of future work would involve studying various new perturbation techniques to further impact the resilience of the object categories. We could also explore multiple new vehicle datasets and explore the effect of perturbations across datasets and algorithms.

\bibliographystyle{IEEEtran}
\bibliography{IEEEabrv,references}

\begin{thebibliography}{10}
\providecommand{\url}[1]{#1}
\csname url@samestyle\endcsname
\providecommand{\newblock}{\relax}
\providecommand{\bibinfo}[2]{#2}
\providecommand{\BIBentrySTDinterwordspacing}{\spaceskip=0pt\relax}
\providecommand{\BIBentryALTinterwordstretchfactor}{4}
\providecommand{\BIBentryALTinterwordspacing}{\spaceskip=\fontdimen2\font plus
\BIBentryALTinterwordstretchfactor\fontdimen3\font minus
  \fontdimen4\font\relax}
\providecommand{\BIBforeignlanguage}[2]{{%
\expandafter\ifx\csname l@#1\endcsname\relax
\typeout{** WARNING: IEEEtran.bst: No hyphenation pattern has been}%
\typeout{** loaded for the language `#1'. Using the pattern for}%
\typeout{** the default language instead.}%
\else
\language=\csname l@#1\endcsname
\fi
#2}}
\providecommand{\BIBdecl}{\relax}
\BIBdecl

\bibitem{biggio2018wild}
B.~Biggio and F.~Roli, ``Wild patterns: Ten years after the rise of adversarial
  machine learning,'' \emph{Pattern Recognition}, vol.~84, pp. 317--331, 2018.

\bibitem{univadpersurvey}
A.~Chaubey, N.~Agrawal, K.~Barnwal, K.~K. Guliani, and P.~Mehta, ``Universal
  adversarial perturbations: A survey,'' 2020.

\bibitem{universaladvpert}
S.-M. Moosavi-Dezfooli, A.~Fawzi, O.~Fawzi, and P.~Frossard, ``Universal
  adversarial perturbations,'' in \emph{Proceedings of the IEEE Conference on
  Computer Vision and Pattern Recognition (CVPR)}, July 2017.

\bibitem{szegedy}
C.~Szegedy, W.~Zaremba, I.~Sutskever, J.~Bruna, D.~Erhan, I.~Goodfellow, and
  R.~Fergus, ``Intriguing properties of neural networks,'' in
  \emph{International Conference on Learning Representations}, 2014.

\bibitem{ozdag}
M.~Ozdag, ``Adversarial attacks and defenses against deep neural networks: A
  survey,'' \emph{Procedia Computer Science}, vol. 140, pp. 152--161, 2018,
  cyber Physical Systems and Deep Learning Chicago, Illinois November 5-7,
  2018.

\bibitem{akhtar}
N.~{Akhtar} and A.~{Mian}, ``Threat of adversarial attacks on deep learning in
  computer vision: A survey,'' \emph{IEEE Access}, vol.~6, pp.
  14\,410--14\,430, 2018.

\bibitem{chakraborty}
A.~Chakraborty, M.~Alam, V.~Dey, A.~Chattopadhyay, and D.~Mukhopadhyay,
  ``Adversarial attacks and defences: {A} survey,'' \emph{CoRR}, vol.
  abs/1810.00069, 2018.

\bibitem{kurakin}
A.~Kurakin, I.~Goodfellow, and S.~Bengio, ``Adversarial machine learning at
  scale,'' 2017.

\bibitem{serban}
A.~{Serban}, E.~{Poll}, and J.~{Visser}, ``{Adversarial Examples on Object
  Recognition: A Comprehensive Survey},'' \emph{arXiv e-prints}, p.
  arXiv:2008.04094, Aug. 2020.

\bibitem{deng}
Y.~{Deng}, X.~{Zheng}, T.~{Zhang}, C.~{Chen}, G.~{Lou}, and M.~{Kim}, ``An
  analysis of adversarial attacks and defenses on autonomous driving models,''
  in \emph{2020 IEEE International Conference on Pervasive Computing and
  Communications (PerCom)}, 2020, pp. 1--10.

\bibitem{sharma}
P.~{Sharma}, D.~{Austin}, and H.~{Liu}, ``Attacks on machine learning:
  Adversarial examples in connected and autonomous vehicles,'' in \emph{2019
  IEEE International Symposium on Technologies for Homeland Security (HST)},
  2019, pp. 1--7.

\bibitem{benz}
P.~{Benz}, C.~{Zhang}, T.~{Imtiaz}, and I.~S. {Kweon}, ``{Double Targeted
  Universal Adversarial Perturbations},'' \emph{arXiv e-prints}, p.
  arXiv:2010.03288, Oct. 2020.

\bibitem{sitawarin}
C.~Sitawarin, A.~Bhagoji, A.~Mosenia, M.~Chiang, and P.~Mittal, ``Darts:
  Deceiving autonomous cars with toxic signs,'' \emph{ArXiv}, vol.
  abs/1802.06430, 2018.

\bibitem{metzen}
J.~H. Metzen, M.~C. Kumar, T.~Brox, and V.~Fischer, ``Universal adversarial
  perturbations against semantic image segmentation,'' in \emph{{IEEE}
  International Conference on Computer Vision, {ICCV} 2017, Venice, Italy,
  October 22-29, 2017}.\hskip 1em plus 0.5em minus 0.4em\relax {IEEE} Computer
  Society, 2017, pp. 2774--2783.

\bibitem{advpertod}
D.~Li, J.~Zhang, and K.~Huang, ``Universal adversarial perturbations against
  object detection,'' \emph{Pattern Recognition}, p. 107584, 2020.

\bibitem{yingwei}
Y.~Li, S.~Bai, C.~Xie, Z.~Liao, X.~Shen, and A.~L. Yuille, ``Regional
  homogeneity: Towards learning transferable universal adversarial
  perturbations against defenses,'' in \emph{Computer Vision - {ECCV} 2020 -
  16th European Conference, Glasgow, UK, August 23-28, 2020, Proceedings, Part
  {XI}}, ser. Lecture Notes in Computer Science, A.~Vedaldi, H.~Bischof,
  T.~Brox, and J.~Frahm, Eds., vol. 12356.\hskip 1em plus 0.5em minus
  0.4em\relax Springer, 2020, pp. 795--813.

\bibitem{fasterRCNN}
S.~Ren, K.~He, R.~Girshick, and J.~Sun, ``Faster r-cnn: Towards real-time
  object detection with region proposal networks,'' in \emph{Advances in Neural
  Information Processing Systems 28}, C.~Cortes, N.~D. Lawrence, D.~D. Lee,
  M.~Sugiyama, and R.~Garnett, Eds.\hskip 1em plus 0.5em minus 0.4em\relax
  Curran Associates, Inc., 2015, pp. 91--99.

\bibitem{detectron2}
Y.~Wu, A.~Kirillov, F.~Massa, W.-Y. Lo, and R.~Girshick, ``Detectron2,'' 2019.

\end{thebibliography}
\end{document}